\documentclass[letterpaper, 10 pt, conference]{ieeeconf}  

\IEEEoverridecommandlockouts                              

\usepackage{algorithm}
\usepackage{algorithmic}
\usepackage{blindtext}
\usepackage{graphicx}
\usepackage[utf8]{inputenc}
\usepackage{mathtools}
\usepackage{amsmath, amsfonts, amssymb,color}
\usepackage{tikz, cite}
\usepackage[font=footnotesize,labelformat=simple]{subcaption}
\usepackage{tabu}

\usepackage{mathrsfs}

\usepackage{multirow}
\usepackage{array,booktabs,arydshln,xcolor}
\usepackage{soul}
\usepackage{mathtools}
\usepackage{nccmath}
\usepackage{caption}
\usepackage{nccmath}
\usepackage{amsmath}
\usepackage{comment}
\usepackage{pdfpages}

\usepackage{float}
\usepackage{graphicx}
\usepackage{epstopdf}
\usepackage{color}
\usepackage{verbatim}
\usepackage{tikz,times}
\usepackage{xcolor}
\usepackage{lipsum}
\usepackage{makecell}

\usepackage{stfloats}

\usepackage{xcolor}

\usepackage{hyperref}
\hypersetup{colorlinks=true, linkcolor=black, citecolor=black, urlcolor=blue}

\title{\LARGE \bf
An Adaptive Transition Framework for Game-Theoretic Based Takeover
}

\author{Dikshant Shehmar$^{1}$, Matthew E. Taylor$^{1}$, Ehsan Hashemi$^{1}$, \IEEEmembership{Senior Member, IEEE} 
\thanks{$^{1}$ Dikshant Shehmar and Matthew E. Taylor (Director of the Intelligent Robot Learning Lab) are with the Computing Science Department, and Ehsan Hashemi (\textit{Corresponding author}, and the Director of the NODE lab) is with the Department of Mechanical Engineering, University of Alberta, Edmonton, AB, Canada, {\tt\small ehashemi@ualberta.ca}. This work is supported by the Natural Science and Engineering Research Council of Canada, and New Frontiers in Research Fund, NFRFE-2022-00795.}
}

\begin{document}
\maketitle
\bstctlcite{BSTcontrol}
\thispagestyle{empty}
\pagestyle{empty}

\begin{abstract}
The transition of control from autonomous systems to human drivers is critical in automated driving systems, particularly due to the out-of-the-loop (OOTL) circumstances that reduce driver readiness and increase reaction times. Existing takeover strategies are based on fixed time-based transitions, which fail to account for real-time driver performance variations. This paper proposes an adaptive transition strategy that dynamically adjusts the control authority based on both the time and tracking ability of the driver trajectory. Shared control is modeled as a cooperative differential game, where control authority is modulated through time-varying objective functions instead of blending control torques directly. To ensure a more natural takeover, a driver-specific state-tracking matrix is introduced, allowing the transition to align with individual control preferences. Multiple transition strategies are evaluated using a cumulative trajectory error metric. Human-in-the-loop control scenarios of the standardized ISO lane change maneuvers demonstrate that adaptive transitions reduce trajectory deviations and driver control effort compared to conventional strategies. Experiments also confirm that continuously adjusting control authority based on real-time deviations enhances vehicle stability while reducing driver effort during takeover.
\end{abstract}

\section{Introduction} \label{sec:Intro}

The evolution of autonomous vehicles promises improved safety, convenience and accessibility in transportation, yet it also presents new challenges in maintaining safe and effective interactions between human drivers and autonomous
systems The Society of Automotive Engineers (SAE) \cite{SAE} defines five levels of autonomy, where intermediate levels (3 and 4) require human drivers to be ready to take control when necessary. However, research shows that humans struggle to maintain attention in highly automated systems \cite{stanton} and often over-rely on automation, leading to the Out-Of-The-Loop (OOTL) phenomenon \cite{endsley1995out}, where drivers fail to properly monitor the system.  

Two key responsibilities for humans in automation are monitoring system performance and being prepared to resume control when automation fails to meet expectations \cite{bainbridge1983ironies}. Studies \cite{stanton2005driver,comfort} have shown that increasing automation reduces mental workload and situational awareness while increasing driver reaction times \cite{young2007back}. These effects are particularly concerning when drivers need to quickly take over control during unexpected events.  

The process of transferring control from automation back to the human driver is called a \textit{takeover} \cite{drexler}. During this transition, control must be smoothly transferred to prevent instability. The National Highway Traffic Safety Administration (NHTSA) \cite{laneChange} has documented numerous accidents caused by unintended lane changes during control transitions. Although Advanced Driver Assistance Systems (ADAS) improve lane-keeping performance, sudden control shifts remain a risk. Instead of an abrupt takeover, shared control -- where both driver and automation contribute simultaneously -- has been shown to improve safety and stability \cite{frank}. One practical implementation of shared control that has gained support \cite{shared,Mosbach,nguyen} involves both the driver and the autonomous system applying the steering torque simultaneously \cite{torque}. 

\begin{figure}[H]
    \centering
    \includegraphics[width=0.9\linewidth]{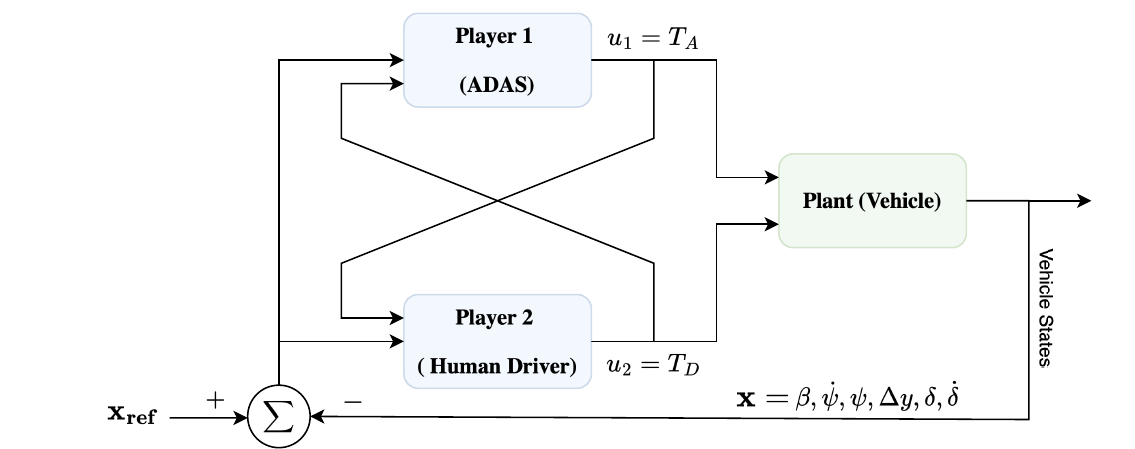}
    \caption{Game theoretic realization for shared control}
    \label{fig:share}
\end{figure}

Game theory provides a mathematical framework for modeling shared control, capturing real-time interactions between the driver and ADAS as a differential game \cite{gametheory}. The closed-loop Nash equilibrium approach has been widely applied in vehicle control, demonstrating robustness in handling uncertainties and adapting to driver behavior \cite{tamaddoni,prof}. Prior work has
formalized shared control as a differential game, ensuring that neither the driver nor the ADAS can unilaterally improve performance by modifying their steering torque inputs, resulting in an equilibrium in steering control \cite{conditions}. This formulation enables precise modeling of mutual influences between both control partners. They further established fundamental design principles for shared control systems, including necessary and sufficient conditions for controller synthesis. As illustrated in Fig.~\ref{fig:share}, their approach conceptualizes the assistance system as a complementary partner to the human drive. This interaction is mathematically formulated as a differential game, enabling precise modeling of mutual influences between both control partners.

Ludwig et al. \cite{Ludwig} focused on developing a mathematical framework for smooth transitions of control authority during takeover scenarios. Their approach, validated through lane change simulations, explored various time-varying takeover strategies. However, their investigation was limited to time-dependent transitions. Our work extends their foundation by considering transition functions that incorporate situational awareness, driver-specific preferences, and quantitative evaluation methods to enhance the safety and efficiency of control transitions. The key contributions of this paper are as follows:
\begin{itemize}
    \item Development of an adaptive transition function that adjusts the control authority based on both time and driver performance.
    \item Comparative analysis of multiple transition strategies using a cumulative error metric for trajectory deviation and stabilization.
    \item Integration of driver-specific control preferences into the shared control framework to ensure efficient takeover.
\end{itemize}

The remainder of this paper is organized as follows. Section \ref{sec:model} presents the vehicle dynamics model and the driver modeling approach. Section \ref{sec:framework} develops the cooperative game-theoretic framework for shared control, establishing the mathematical foundations for driver-ADAS interaction. Section \ref{sec:sharing} analyzes various control sharing strategies, including the proposed situation-aware transition functions and the integration of driver preference. Section \ref{sec:results} validates these approaches through simulation studies of lane-change and overtaking scenarios, with quantitative performance comparisons. Finally, Section \ref{sec:conclusion} summarizes the key findings and discusses future research directions.

\section{Model Description} \label{sec:model}
\subsection{Shared Control Steering Framework}
In the shared control setup, both the human driver and the ADAS apply torques to the steering wheel. Instead of explicitly blending their control inputs as a convex combination, the system dynamically adjusts control authority by modifying the weighting matrices \(Q_D\) (driver) and \(Q_A\) (ADAS) in the optimal control formulation \cite{Ludwig}. Formally, the cost function weights are adjusted as follows: 
\begin{align}
    Q_D(t) = \alpha(t)Q_D^{\max}, \quad Q_A(t) = (1-\alpha(t))Q_A^{\max}  
\end{align}
for some function \(\alpha(t)\). More details about this update rule and transition function is provided in Section \ref{sec:sharing}.
The shared control steering system allows for interaction between the human driver and the ADAS through the steering wheel. Furthermore, both the driver and the ADAS have access to the vehicle states and are guided by predefined reference trajectories.  This interaction facilitates implicit communication, where the driver can sense ADAS inputs through steering feedback, improving coordination during takeover.

\subsection{Vehicle Model}
\begin{figure}
    \centering
    \includegraphics[width=0.8\linewidth]{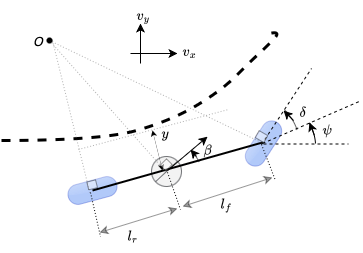}
    \caption{Single Track Vehicle Model}
    \label{fig:bicycle_model}
\end{figure}

We consider an extended single-track linear vehicle model with an additional steering wheel dynamics model (see Fig. \ref{fig:bicycle_model}). The single-track model is a simplified representation of a vehicle, where the two front and rear wheels are combined into single equivalent wheels located at the center of each axle. The steering wheel dynamics model represents the lateral and yaw dynamics of the vehicle \cite{sergey2000}, providing a good approximation for steering maneuvers with lateral accelerations up to \(4 \, \text{m/s}^2\).

Assuming that the speed of vehicle \(v\) remains constant during the analysis, the dynamics of the vehicle and the steering wheel is described by the following time-invariant linear state-space model as follows:

\begin{equation}
    \mathbf{\dot{x}}(t) = \mathbf{A}\mathbf{x}(t) + \mathbf{B_D}T_D(t) + \mathbf{B_A}T_A(t),
    \label{eqn:state}
\end{equation}

\noindent
where \(\mathbf{x}(t)\) is the state vector, \(\mathbf{A}\) is the vehicle system matrix, and \(\mathbf{B_D}, \mathbf{B_A}\) are the input weight vectors, defined in Appendix \ref{sec:sys}. 
The state vector \(\mathbf{x}(t)\) is defined as: $\mathbf{x}(t) = [\beta(t) \quad \dot{\psi}(t) \quad \psi(t) \quad  y(t) \quad  \delta(t) \quad \dot{\delta}(t)]^\top,$ where, \(\beta(t)\) is the side-slip angle of the vehicle, defined as \(\beta = \tan^{-1}(v_y / v_x)\), where \(v_x\) and \(v_y\) are the components of longitudinal and lateral velocity, respectively, \(\psi(t)\) represents the vehicle yaw angle, \(\dot{\psi}(t)\) is the yaw rate, \(y(t)\) represents the vehicle lateral displacement from the center of the driving lane, \(\delta(t)\) is the steering wheel angle, and \(\dot{\delta}(t)\) is the rate of change of the steering. \(\delta(t)\) is modeled as linearly related to the steering angle \(\delta_s(t)\) by \(\delta(t) = i_s\delta_s(t)\). Here, \(T_D(t)\) and \(T_A(t)\) are the torques applied by the driver and the assistance system, respectively, on the steering wheel.

\section{Control Framework} \label{sec:framework}

This section presents the cooperative game-theoretic framework utilized to solve the problem. Cooperative game theory provides a mathematical basis where all players have access to each other's cost functionals. As mentioned in Section \ref{sec:model}, both the driver and the ADAS have access to each other's states. The focus is on group behavior, emphasizing coalitions between players to achieve collective goals \cite{book}.

\subsection{General Formulation}

In a \textit{cooperative differential game} setting, there are $N$ players denoted by the set $\mathscr{P}=\{1, 2, \ldots, N\}$, who interact via a continuous-time dynamical system. The system is governed by first-order ordinary differential equations over a time horizon $T > 0$ which can be described as:
\begin{equation}
    \dot{x}(t) = f(t, x(t), u_1(t), \ldots, u_N(t)), \quad x(0) = x_0 \in \mathbb{R}^n, \label{eqn:dynamics}
\end{equation}

\noindent
where, $x(t) \in \mathbb{R}^n$: state vector of the game at time $t$, $x_0$: initial state of the game,  $u_i(t) \in \mathcal{U}_i$: control input of player $i$ from the admissible set $\mathcal{U}_i \subset \mathbb{R}^{m^i}$,  $f: [0,T] \times \mathbb{R}^n \times \mathcal{U}_1 \times \ldots \times \mathcal{U}_N \to \mathbb{R}^n$: governing function (potentially non-linear). Each player $i \in \mathscr{P}$ aims to optimize their performance by minimizing the following cost functional:
\begin{equation}
    J_i(x, u_1, .., u_N) \triangleq \int_0^T g_i(t, x(t), u_1(t), \ldots, u_N(t)) dt, 
    \label{eqn:costFunction}
\end{equation}
where $g_i: [0,T] \times \mathbb{R}^n \times \mathcal{U}_1 \times \cdots \times \mathcal{U}_N \to \mathbb{R}$ is the instantaneous or running cost rate for player $i$. Note that each player's control decisions affect not only their own cost, but also the outcomes for other players. 

The information structure defines the knowledge available to players when making decisions. There are two primary paradigms, Open and Closed Loop Information Structures. In \textit{Open-Loop Information Structure} players only know the initial state $x_0$ of the game and the control strategy $\gamma_i$, a mapping  $\mathbb{R}^n \to \mathcal{U}_i$ from the initial state to control actions, such that $u_i(t) = \gamma_i(t, x_0) \forall t \in [0, T]$. In \textit{Closed-Loop Information Structure}, players have access to the current state $x(t)$ of the game at all times and the control strategy, $\phi_i: \mathbb{R}^n \to \mathcal{U}_i$, that maps the current state and time to control actions, such that $u_i(t) = \phi_i(t, x(t)) \forall t \in [0, T]$. In this work, we adopt the closed-loop information structure, as it allows players to use real-time information for decision making, improving responsiveness and adaptability.

A \textit{Nash equilibrium} is a collection of strategies in which no player has an incentive to unilaterally deviate from their chosen strategy. In the cooperative dynamic game context, Nash equilibrium ensures that all players act optimally with respect to each other's strategies. Explicitly, the Nash equilibrium for two-player game can be defined by the set of strategies $\{u_{i}^*(t)\}_{i=1}^2$ that satisfy:
\begin{equation}
\begin{split}
    J^*_1 \triangleq J_1(t,  x, u^*_1(t), u^*_2(t)) \leq J_1(t, x, u_1(t), u^*_2(t)) \\
    J^*_2 \triangleq J_2(t, x, u^*_1(t), u^*_2(t)) \leq J_2(t, x, u^*_1(t), u_2(t))
    \label{eqn:nash_equilibrium}
\end{split}
\end{equation}
$\{\forall u_i(t), i = 1, 2\}$ subject to the dynamics in \eqref{eqn:dynamics} and the admissible control sets $\mathcal{U}_i$. For a Nash equilibrium to exist, the cost functionals $J_i$ must be continuously differentiable with respect to $u_i$, the admissible control sets $\mathcal{U}_i$ must be convex, and the governing function $f(t, x, u_1, u_2)$ should satisfy the Lipschitz continuity in $x$ and $u_i$ to ensure the existence of a unique solution to differential equations.

\subsection{Cooperative Differential Game Setup for Takeover}

We consider a cooperative differential game involving two players working towards a shared goal of tracking a reference path smoothly. The two players are the \textbf{human driver}, exerting a control torque $T_D$ and the \textbf{ADAS}, applying a control torque $T_A$.
Let $\mathbf{x}_{\text{ref}}(t)$ denote the reference state, then the tracking error can be defined as $\tilde{\mathbf{x}} = \mathbf{x}(t) - \mathbf{x}_{\text{ref}}$. We can modify the dynamics of the system \eqref{eqn:state} to incorporate tracking error dynamics using:
\begin{equation}
    \dot{\tilde{\mathbf{x}}}(t) = \mathbf{A}\tilde{\mathbf{x}}(t) + \mathbf{B_D}T_D(t) + \mathbf{B_A}T_A(t) + \mathbf{A}\mathbf{x}_{\text{ref}}
\end{equation}

The players aim to minimize their respective cost functionals, which are quadratic in nature. Quadratic cost formulations are widely used in optimal control and game-theoretic settings due to their convexity and analytical tractability \cite{quadraticcostfunction}. From \eqref{eqn:costFunction}, the cost functional for a player $i \in \{1, 2\}$ can be defined as:
\begin{equation}
    \begin{split}
        J_i = \frac{1}{2}\tilde{\mathbf{x}}^\top(T) \mathbf{S_i} \tilde{\mathbf{x}}(T) 
        + \frac{1}{2}\int_0^T \big\{\tilde{\mathbf{x}}^\top(t)\mathbf{Q_i}(t)\tilde{\mathbf{x}}(t) \\
        + \mathbf{u}_i^\top(t)\mathbf{R_i}(t)\mathbf{u}_i(t)\big\} dt
    \end{split}
\label{eqn:cost}
\end{equation}

\noindent
where, $\mathbf{S_i}$ is the terminal state weighting matrix (symmetric and positive semidefinite), $\mathbf{Q_i}(t)$ represents the state deviation penalty matrix (symmetric and positive semidefinite), and $\mathbf{R_i}(t)$ represents the control effort penalty matrix (positive definite). The choice of this quadratic structure ensures a trade-off between stability and performance, preventing excessive control actions while guiding the system towards the desired state. The objective for each player $i$ is to solve the linear-quadratic optimal control problem:
\begin{equation}
        \mathbf{u}_i^*(\cdot) = \arg\min_{\mathbf{u}_i(\cdot)} J_i \\
\end{equation}
subject to the dynamics:
\begin{equation}
    \begin{aligned}
        \dot{\tilde{\mathbf{x}}}(t) &= \mathbf{A} \tilde{\mathbf{x}}(t) + \mathbf{B}_i \mathbf{u}_i(t) + \mathbf{B}_j \mathbf{u}_j^*(t) + \mathbf{A} \mathbf{x}_{\text{ref}}, j \neq i, \\
        \tilde{\mathbf{x}}(0) &= \mathbf{x}_0 - \mathbf{x}_{\text{ref}}
    \end{aligned}
\end{equation}

We assume that the final cost function, after solving \eqref{eqn:cost}, takes the quadratic form: 
\begin{equation}
    V_i(\tilde{\mathbf{x}}, t) = \frac{1}{2} \tilde{\mathbf{x}}^\top \mathbf{P_i}(t) \tilde{\mathbf{x}} + c_i(t)
    \label{eqn:final_val}
\end{equation}

\noindent
where $\mathbf{P_i}(t)$ is symmetric positive semi-definite matrix and $c_i(t)$ is a scalar function. The optimal control strategies $\mathbf{u}_1^*$ and $\mathbf{u}_2^*$ are derived using the Nash equilibrium condition by solving coupled Hamilton-Jacobi-Bellman equation (HJB) ($0 = J_t^*(\mathbf{x}(t), t) + \mathscr{H}(\mathbf{x}(t), \mathbf{u}^*(\mathbf{x}(t), J_\mathbf{x}^*, t), J_\mathbf{x}^*, t)$) \cite{Kirk1970OptimalCT}. Differentiating the Hamiltonian with respect to $\mathbf{u}_i$, we obtain:
\begin{equation}
\begin{split}
    \mathbf{u}_i^* &= -\mathbf{R_i}^{-1} \mathbf{B}_i^\top \nabla_{\tilde{\mathbf{x}}} V_i = -\mathbf{R_i}^{-1} \mathbf{B}_i^\top \big( \mathbf{P_i}(t)\tilde{\mathbf{x}}\big).
\end{split}
\label{eqn:optimalcontrol}
\end{equation}

\noindent
with boundary condition, \( \nabla_{\tilde{\mathbf{x}}} V_i (\tilde{\mathbf{x}}, T) = \mathbf{S_i}\tilde{\mathbf{x}}(T)\). 
Now, we have \( V_i (\tilde{\mathbf{x}}, T) = \frac{1}{2}\tilde{\mathbf{x}}(T)^\top \mathbf{S_i}\tilde{\mathbf{x}}(T)\) from \eqref{eqn:cost}
 and \( V_i(\tilde{\mathbf{x}}, T) = \frac{1}{2} \tilde{\mathbf{x}}^\top \mathbf{P_i}(T) \tilde{\mathbf{x}} + c_i(T)\) from \eqref{eqn:final_val}. Comparing these two, we get the boundary condition \(\mathbf{P_i}(T)=\mathbf{S_i}, \text{ and } c_i(T)=0\). We also have $\nabla_\mathbf{t}V_i = \frac{1}{2} \tilde{\mathbf{x}}^\top \dot{\mathbf{P_i}}(t) \tilde{\mathbf{x}} + \dot{c_i}(t) $ and $\nabla_{\tilde{\mathbf{x}}} V_i = \big(\mathbf{P_i}(t)\tilde{\mathbf{x}} \big){^\top}$. Substituting back into the HJB equations, we will get
\begin{equation}
\begin{split}
& \left( \frac{1}{2} \tilde{\mathbf{x}}^\top \mathbf{\dot{P}}_i(t) \tilde{\mathbf{x}} + \dot{c}_i(t) \right) = \\
& - \Biggl\{ \bigg(\mathbf{P_i}(t)\tilde{\mathbf{x}} \bigg){^\top} \bigg( \mathbf{A} \tilde{\mathbf{x}}(t) + \sum_{j=1}^{2} \mathbf{B_j} \mathbf{u_j^*}(t) + \mathbf{A} \mathbf{x}_\text{ref} \bigg)  \\
& + \frac{1}{2} \tilde{\mathbf{x}}^\top(t)\mathbf{Q_i}(t)\tilde{\mathbf{x}}(t)
+ \frac{1}{2} \mathbf{u}_i^*{^\top(t)}\mathbf{R_i}(t)\mathbf{u}_i^*(t) \Biggl\}  \label{eqn:final}
\end{split}
\end{equation}

Collecting quadratic terms from \eqref{eqn:final}, we will get the Riccati Differential equation for $\mathbf{P}$: 
\begin{equation}
\begin{split}
\mathbf{\dot{P}}_i(t) = -\Big( \mathbf{P_i}(t)\mathbf{A} + \mathbf{A}^\top \mathbf{P_i}(t) - \mathbf{P_j}(t)^\top \mathbf{B_j^\top R_j^{-1}B_j} \mathbf{P_i}(t) \\
 - \mathbf{P_i}(t) \mathbf{B_j R_j^{-1}B_j^\top} \mathbf{P_j}(t) - \mathbf{P_i}(t) \mathbf{B_i R_i^{-1}B_i^\top} \mathbf{P_i}(t) + \mathbf{Q_i}  \Big) \\
\end{split}
\label{eqn:riccati}
\end{equation}
for $j \neq i$, with boundary conditions defined as: $\mathbf{P_i}(T) = \mathbf{S_i}$. After obtaining this $\mathbf{P_i}$ recursively, we can get the optimal control for both players by substituting it in \eqref{eqn:optimalcontrol}. Since (\ref{eqn:riccati}) do not have closed-form analytical solutions in general, they are typically solved using numerical methods \cite{FREILING1996291}. A common approach is to employ a first-order backward Euler discretization for numerical integration, where the equations are propagated backward in time from a terminal condition. 
Hence, in a discrete-time framework, the optimal control problem is reformulated, with state and control inputs evaluated at discrete steps. The cost function in discrete time is given by:
$J = \sum_{k=0}^{N-1} \left[ \frac{1}{2} \mathbf{x}_k^\top \mathbf{Q} \mathbf{x}_k + \frac{1}{2} \mathbf{u}_k^\top \mathbf{R} \mathbf{u}_k \right] + \frac{1}{2} \mathbf{x}_N^\top \mathbf{S} \mathbf{x}_N$, where $\mathbf{S}$ is the terminal cost weight. The corresponding discrete-time Riccati equations governing the evolution of $\mathbf{P}_k$ are:
\begin{equation}
\begin{split}
    \mathbf{P}_i^{(k-1)} = \mathbf{P}_i^{(k)} + \Big( \mathbf{P}_i^{(k)} (\mathbf{A}_d + \mathbf{F}_j \mathbf{P}_j^{(k)}) \\
    + (\mathbf{A}_d - \mathbf{F}_j \mathbf{P}_j^{(k)})^\top \mathbf{P}_i^{(k)} - \mathbf{P}_i^{(k)} \mathbf{F}_i \mathbf{P}_i^{(k)} + \mathbf{Q}_i \Big) \Delta t,
\end{split}
\end{equation}
for $j \neq i$, where the index $k = n, \ldots, 1$ with $n = T / \Delta t$, $\mathbf{P}_i^{(n)} = \mathbf{S}$, and  $\mathbf{F}_i = \mathbf{B}_i \mathbf{R}_i^{-1} \mathbf{B}_i^\top$.

The stability of the shared control strategy arises from the closed-loop Nash equilibrium, where neither the driver nor the ADAS can unilaterally improve their strategy. This ensures a balanced interaction, preventing destabilizing behaviors. The coupled Riccati equations governing the optimal control laws yield symmetric, positive-definite solutions under standard controllability and observability conditions, ensuring deviations are corrected through feedback control. The existence and uniqueness of the Nash equilibrium further reinforce stability. Thus, the cooperative game-theoretic approach ensures stability through real-time adaptation, closed-loop feedback, and well-structured cost functions. 

\section{Sharing Controlled Strategy} \label{sec:sharing}
Towards the end of the takeover process, the human driver must regain full control of the vehicle. The cooperative differential game framework introduced in Section~\ref{sec:framework} accommodates time-varying matrices, allowing it to model the evolving objectives of both the human driver and the ADAS during the transition. A gradual shift in control authority is necessary to ensure safety and stability, which is implemented by adjusting the weighting matrices \( \mathbf{Q} \) and \( \mathbf{R} \) in the respective cost functions of both agents. To facilitate a smooth transition, the cooperative game framework dynamically modifies these matrices. Prior work \cite{Ludwig} suggests that modulating \( \mathbf{Q} \) is more computationally efficient than adjusting \( \mathbf{R} \), making it the preferred approach.  

\subsection{Control Authority Transition Function}  

The shift in control authority between the ADAS and the human driver is governed by a function \( \alpha(t) \), defined as:  

\begin{equation}
\alpha(t) \in [0,1], \quad \forall t \in [t_S, t_E]
\end{equation}

\noindent
where \( t_S \) and \( t_E \) denote the start and end of the transition, respectively. By definition, \( \alpha(t_S) = 0 \) (full ADAS control) and \( \alpha(t_E) = 1 \) (full human control). At the end of the transition, the driver is expected to have full control of the vehicle. During the transition phase, the tracking matrices for both agents evolve as follows:  

\begin{equation}
\mathbf{Q_D}^{(t)} = \alpha(t)\mathbf{Q_D^{\max}}, \quad
\mathbf{Q_A}^{(t)} = (1 - \alpha(t))\mathbf{Q_A^{\max}}
\end{equation}

\noindent
where \( \mathbf{Q_D^{\max}} \) and \( \mathbf{Q_A^{\max}} \) represent the maximum weighting matrices for the human driver and the ADAS, respectively. This gradual modulation prevents abrupt control shifts that can destabilize the system. Several transition functions can be used to define how control shifts from the ADAS to the human driver during the given takeover time.

(i) The step function provides an instantaneous transition at a specified time point, but this abrupt change can introduce discontinuities in vehicle control. (ii) The linear function ensures a constant rate transition, offering a predictable and uniform control transfer. (iii) The sigmoid function, characterized by its S-shaped curve, enables smoother transitions with gradual initiation and completion, controlled by the parameter \( k \). (iv) The cooperative function maintains the equal sharing of control between the human driver and the ADAS throughout the transition, ensuring a balanced contribution. (v) The exponential function offers a rapid initial transition that gradually slows, balancing responsiveness with smoothness. (vi) Finally, the proposed adaptive function dynamically adjusts control authority based on real-time trajectory deviations, increasing ADAS assistance when the driver deviates from the reference state, as measured by cross-track error (\(\varepsilon_y\)) and heading error (\(\varepsilon_\psi\)). Unlike previous transitions, which do not account for real-time driver performance, this approach ensures system stability even if the driver initially struggles, allowing for a more responsive and adaptive takeover.
Table~\ref{tab:transition_functions} summarizes the mathematical forms of these transition functions.  

\begin{table}[htbp]
\centering
\caption{Control Authority Transition Functions}
\renewcommand{\arraystretch}{1.5}
\begin{tabular}{p{2.5cm} p{5.2cm}}
\hline
\textbf{Transition Function} & \textbf{Mathematical Form} \\
\hline
Step & \( \alpha(t) =  
\begin{cases} 0, & t < t_S \\ 1, & t \geq t_S \end{cases} \) \\  
Linear & \( \alpha(t) = \frac{t - t_S}{t_E - t_S} \) \\  
Cooperative & \( \alpha(t) = 0.5 \) \\  
Sigmoid & \( \alpha(t) = 1 - \frac{1}{1 + e^{-k(t - \frac{t_E + t_S}{2(t_E-t_S)})}} \) \\  
Exponential & \( \alpha(t) = 1 - e^{\frac{-\lambda(t-t_S)}{t_E-t_S}}, \quad \lambda>1\) \\  
Adaptive \textbf{[OURS]} & \( \alpha(t) = 1- \min(0.5+|k_1\varepsilon_y + k_2\varepsilon_\psi|, 1.0) \) \\  
\hline
\end{tabular}
\label{tab:transition_functions}
\end{table}  

\subsection{Quantitative Evaluation of Transition Functions}\label{sec:eval}

To assess the effectiveness of different transition strategies, we evaluate their impact on vehicle stability, tracking performance, and driver workload. The primary metric used for comparison is the cumulative trajectory error, defined as:

\begin{equation}
\varepsilon_{total} = \sum_0^T\left(\varepsilon_y^2 + \varepsilon_\psi^2 + \beta^2 + \delta_D^2\right ) 
\label{eqn:error}
\end{equation}

\noindent 
where \( \varepsilon_y \) represents the cross-track error, measuring lateral deviation from the reference trajectory, and \( \varepsilon_\psi \) represents the heading error, indicating misalignment between the vehicle’s orientation and the reference path. The slip angle \( \beta \) is included as it captures vehicle stability during the transition, ensuring that control transfer does not induce excessive lateral forces. The steering angle \( \delta \) reflects the effort required from the driver to maintain the trajectory, while the steering rate \( \dot{\delta} \) serves as an indicator of cognitive load, with rapid steering changes suggesting increased mental effort \cite{comparison}.  

Since drivers require time to stabilize even after full control transfer, the total error is computed over the entire simulation duration \( T \) rather than stopping at \( t_E \). This approach ensures that lingering effects of different transition strategies are captured, particularly how long it takes the driver to align with the reference trajectory post-transition. To enable fair comparisons across transition functions, each error term is normalized before summation, ensuring that the cumulative errors have the same upper-bound. The comparative results for each transition function are presented in Section~\ref{sec:results}.  

\subsection{Human Driver Tracking Preferences}\label{sec:reg}  

Since drivers exhibit different tracking behaviors, an effective transition strategy should account for individual preferences. Some drivers prioritize minimizing cross-track deviation, while others focus on reducing heading error. To incorporate these variations, we estimate individualized weighting matrices \( \mathbf{Q_D} \) based on empirical driver data. 

\begin{figure}
    \centering
    \includegraphics[width=0.75\linewidth, height = 4.7cm]{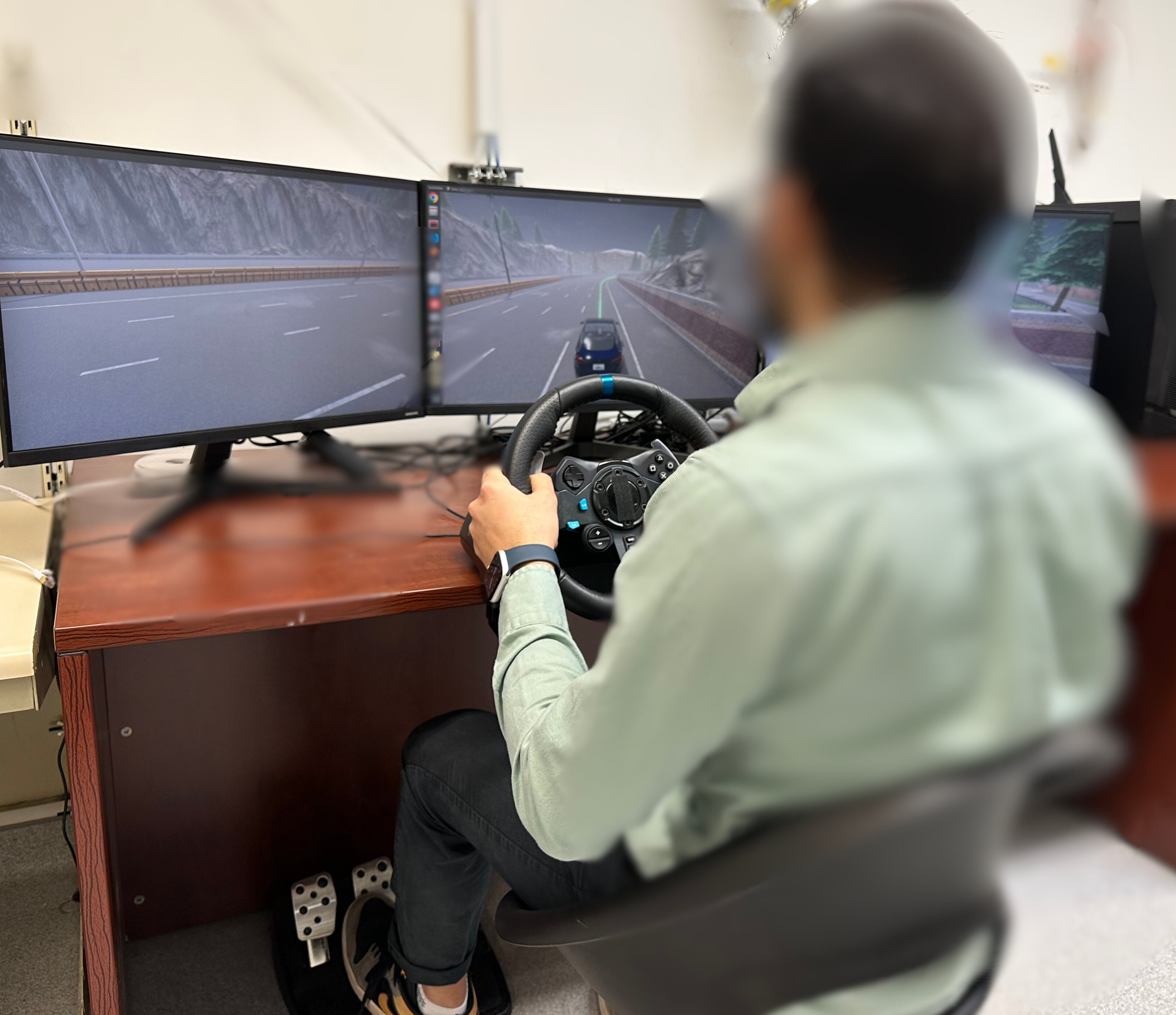}
    \caption{Simulator setup}
    \label{fig:study}
\end{figure}

\begin{algorithm}
\caption{Shared Control Takeover}\label{algo}
\begin{algorithmic}[1]
\REQUIRE
    \STATE Vehicle state: $\mathbf{x} = [\beta, \dot{\psi}, \psi, y, \delta, \dot{\delta}]$
    \STATE Reference trajectory: $\mathbf{x_{\text{ref}}}$
    \STATE Transition time: start $t_S$, end $t_E$
    \STATE Initialize control weights: $(\mathbf{Q_A^{\max}}, \mathbf{Q_D^{\max}}, \mathbf{R_A}, \mathbf{R_D})$
\ENSURE
    \STATE Control signals: $(u_A, u_D)$
    \STATE Authority weights: $(\alpha_A, \alpha_D)$
\FOR{each timestep $t$}
    \IF{$t < t_S$}
        \STATE $\alpha_A \gets 1$, $\alpha_D \gets 0$
    \ELSIF{$t_S \leq t < t_E$}
        \STATE $\alpha_A \gets \alpha_{type}(t)$ \hfill {\color{gray}\textit{// Transition function}}
        \STATE $\alpha_D \gets 1 - \alpha_A$
    \ELSE
        \STATE $\alpha_A \gets 0$, $\alpha_D \gets 1$
    \ENDIF
    
    \STATE $\mathbf{Q_A} \gets \alpha_A \mathbf{Q_A^{\max}}$ \hfill {\color{gray}\textit{// Update weights}}
    \STATE $\mathbf{Q_D} \gets \alpha_D \mathbf{Q_D^{\max}}$
    \STATE $\mathbf{P_A}, \mathbf{P_D} \gets \text{SolveRiccati}(\mathbf{A}, \mathbf{B_{A, D}}, \mathbf{Q_{A,D}}, \mathbf{R_{A,D}})$
    
    \STATE $\tilde{\mathbf{x}} \gets \mathbf{x} - \mathbf{x}_{\text{ref}}$ 
    \STATE $u_D \gets -\mathbf{R_D}^{-1}\mathbf{B_D}^\top \mathbf{P_D}\tilde{\mathbf{x}}$
    \STATE $u_A \gets -\mathbf{R_A}^{-1}\mathbf{B_A}^\top \mathbf{P_A}\tilde{\mathbf{x}}$ \hfill {\color{gray}\textit{// Compute controls}}
    
    \STATE $\mathbf{x}_{t+1} \gets \mathbf{A} \mathbf{x}_t + \mathbf{B_A} u_A + \mathbf{B_D} u_D$ \hfill {\color{gray}\textit{// Update state}}
\ENDFOR
\end{algorithmic}
\end{algorithm}

We conducted an \textit{ethics-approved} human-in-the-loop study (Fig. \ref{fig:study}) with ten licensed drivers (age 23-32, driving experience 1-7 years) in a simulated driving environment. The CARLA simulator was used to replicate realistic vehicle dynamics, and a Logitech G29 steering wheel was employed for manual control. Each driver completed two familiarization runs followed by two recorded test runs, where they were instructed to follow a predefined trajectory. Data were sampled at a fixed interval of \( 0.1s \), capturing vehicle state and control inputs. To maintain consistency, data were recorded only when the vehicle reached a steady velocity, since our system assumes a constant speed. To derive individualized control weights, we formulated the following optimization problem:

\begin{equation}
J_i = \sum_{k=1}^{N} (\mathbf{x}_k - \mathbf{x}_{\text{ref},k})^\top \mathbf{Q_i} (\mathbf{x}_k - \mathbf{x}_{\text{ref},k}) + \mathbf{u}^\top\mathbf{R_i}\mathbf{u}
\label{eqn:reg}
\end{equation}

 \noindent
where \( \mathbf{x}_k \) represents the vehicle state at timestep \( k \), and \( \mathbf{x}_{\text{ref},k} \) is the reference state. The weighting matrix \( \mathbf{Q_i} \) captures driver-specific tracking behavior, while \( \mathbf{R_i} \) regulates control effort \(\mathbf{u}\), which was fixed at 1.0 across all drivers to maintain consistency. The optimal \( \mathbf{Q_i} \) matrix was estimated using regression, ensuring that it remained positive definite.  

By integrating individualized weighting matrices into the transition strategy, control authority is adjusted to match each driver’s natural control style. This ensures that by the time full control is transferred, the driver is already in a preferred control state, minimizing the need for corrective actions. To account for any residual transition effects, the cumulative error is computed over the entire duration of the simulation \(\eqref{eqn:error}\). This approach results in a smoother takeover, enhancing overall stability and driver comfort.

\section{Experimental results and discussion}\label{sec:results}
After collecting human driving data, we performed optimization \eqref{eqn:reg} to obtain \(\mathbf{Q^{max}_D}\) for each driver. The ADAS system was controlled using an LQR controller, where the quadratic control effort penalty was set to \(1\), and the lateral position deviation penalty was set to \(5\), with other states not considered for ADAS control. Once \(\mathbf{Q^{max}}\) was determined for both the ADAS and the human driver, we applied the shared control framework using Algorithm \ref{algo}. To evaluate our approach and compare different transition strategies, we performed simulations following the ISO 17361:2017 (lane change) and ISO 3888-1:2018 (double lane change) standards. In the lane change scenario, the vehicle, traveling at \(120km/h\), must perform a maneuver to avoid an obstacle, with a time-to-collision of \(7 s\). The transition process begins at \(3s\) and ends at \(8s\), with the reference lateral position remaining at \(0m\) until the start of the lane change, where it increases linearly to \(3.75m\). The double lane change scenario introduces greater complexity, requiring the vehicle to change lanes to avoid an obstacle and then return to its original lane. This maneuver demands higher control effort to maintain stability while executing both lane changes. The transition strategy was evaluated with a preview horizon of \(T=1.5s\), optimizing control inputs at every \(0.01s\) discrete step. Each simulation run lasted \(10 s\), ensuring sufficient time to assess the impact of different transition strategies.

\subsection{Comparison of Transition Strategies}  
\begin{figure}[htbp]
    \centering
    \includegraphics[width=0.9\linewidth]{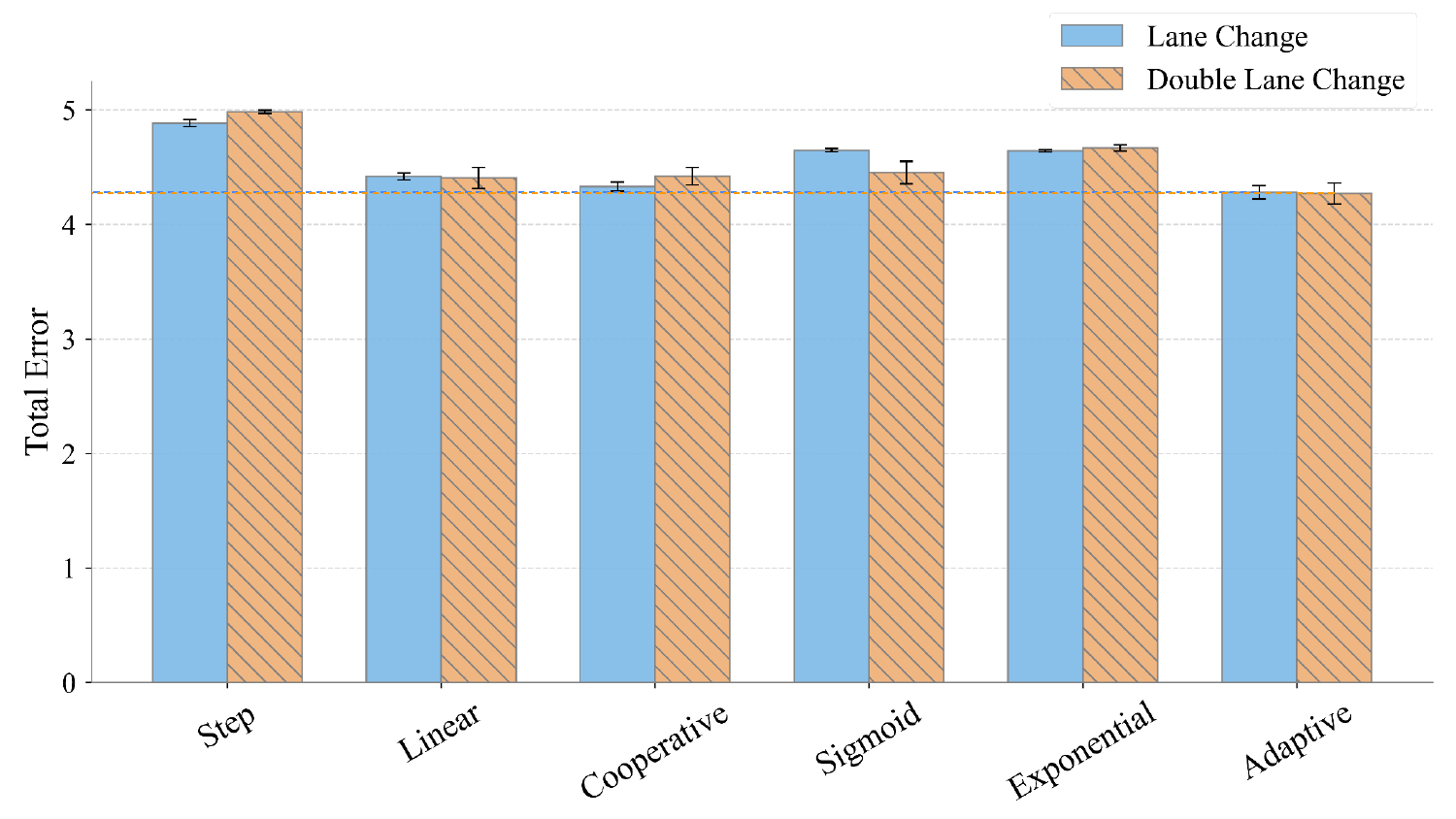}
    \caption{Cumulative error with different transitions}
    \label{fig:final}
\end{figure}

We evaluated all transition strategies listed in Table \ref{tab:transition_functions} using the cumulative trajectory error metric \eqref{eqn:error} by performing simulations as mentioned in previous section. Fig. \ref{fig:final} presents the cumulative error for each transition strategy, with error bars representing the standard deviation across different drivers. The step transition resulted in the highest error in both scenarios, as the ADAS does not provide assistance during the transition, requiring the human driver to assume full control immediately. This led to large deviations from the reference trajectory and an overall unstable transition. The cooperative transition, where control is shared equally between the ADAS and the human driver, performed significantly better, reducing cumulative error in the double lane change scenario by 10.64\% compared to the step transition. However, the adaptive transition achieved the lowest total error, further reducing the error by 14.44\%, demonstrating its effectiveness in dynamically adjusting control authority based on real-time trajectory deviations. By continuously adapting to the driver's performance, the adaptive strategy ensures that when the driver struggles to maintain control, the ADAS provides targeted assistance to stabilize the vehicle. This results in smoother, more controlled transitions, improving both vehicle stability and driver performance.

\subsection{Steering Effort Analysis}  

\begin{figure}[htp]
    \centering
    \includegraphics[width=0.95\linewidth]{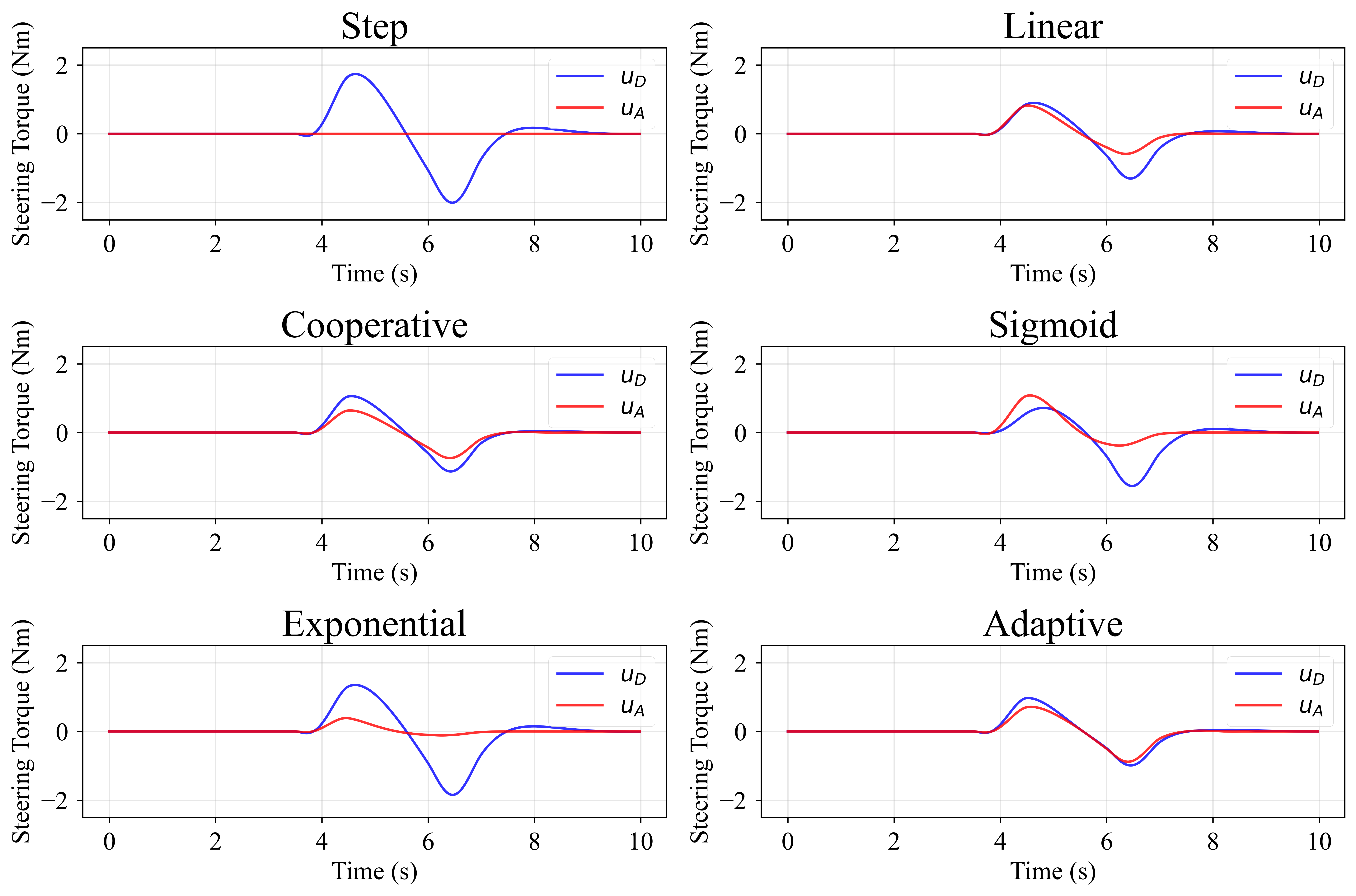}
    \caption{Control input contribution during lane change}
    \label{fig:lc-combined}
\end{figure}

To further examine the impact of different transitions on driver control effort, we analyzed steering input variations for a single participant in the lane change scenario. Fig. \ref{fig:lc-combined} shows the steering input across different transition strategies. In the step transition, the driver’s steering input fluctuates between \(-2.0\) and \(+1.9\), requiring large corrections to maintain the trajectory. In contrast, with adaptive transition, the input range is substantially reduced to \([-0.9, +0.9]\), indicating a smoother and more controlled response.  

\begin{figure}[htp]
    \centering
    \includegraphics[width=0.95\linewidth]{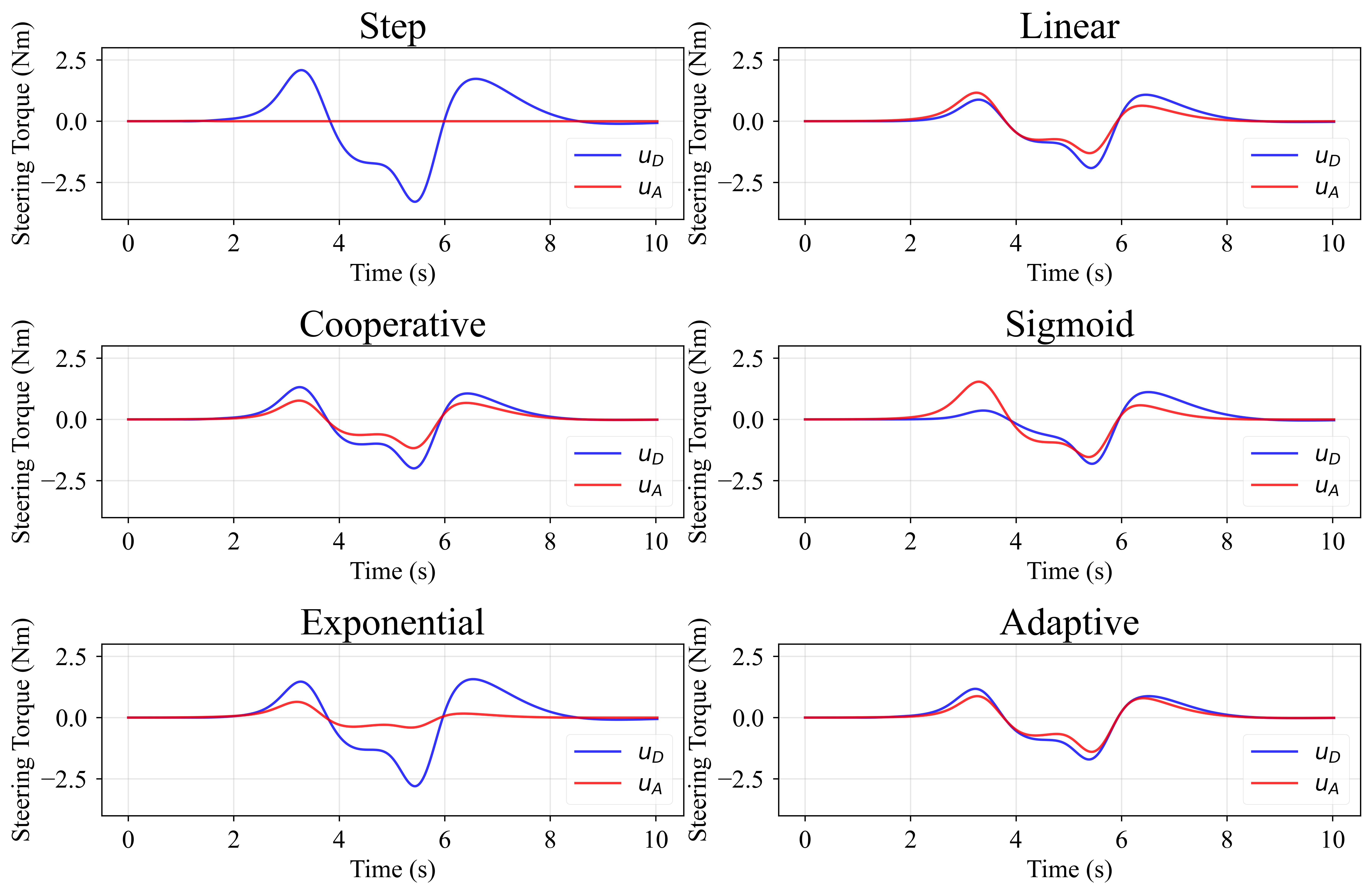}
    \caption{Control input contribution during double lane change}
    \label{fig:o-combined}
\end{figure}

This effect is even more pronounced in the double-lane change scenario, which is inherently more demanding. As shown in Fig. \ref{fig:o-combined}, the driver’s steering input for the step transition varies between \(-3.2\) and \(+2.1\), requiring extensive corrections. However, in the adaptive transition, the range is reduced to \([-1.8, +1.0]\), demonstrating that adaptive control substantially reduces the driver workload and stabilizes the vehicle dynamics.  
The cooperative transition also performs well, showing a balanced distribution of control effort. However, the adaptive approach dynamically reallocates control authority based on real-time driving performance, providing a effective transition. These results highlight the advantages of adaptive transition strategies in reducing driver workload, improving trajectory tracking, and ensuring a more stable takeover process.   

\section{Conclusions}\label{sec:conclusion}

In this paper, we proposed an adaptive transition strategy for shared control during takeovers and introduced a systematic method to compare different transition strategies based on vehicle tracking performance and stability. Using a game-theoretic framework, we dynamically adjusted control authority through time-varying weighting matrices while incorporating real-time driver performance. Furthermore, we integrated driver-specific preferences to ensure that, at the end of the transition, control is transferred in a manner that aligns with the natural behavior of the driver, minimizing the need for corrective actions. The proposed approach was evaluated through simulations based on ISO lane change standards, demonstrating that adaptive transitions improve tracking and stability compared to conventional methods.

Future work will focus on validating this adaptive transition strategy in human-in-the-loop experiments using the CARLA simulator, where human drivers will directly experience shared control during takeovers in real time. The control authority will be transferred dynamically based on real-time driver inputs, allowing for an in-depth analysis of transition performance across different human drivers.

\bibliographystyle{IEEEtran}
\bibliography{main}

\appendices
\section{Vehicle Model} \label{sec:sys}
The system matrix \(\mathbf{A}\) is given by:
\[
\mathbf{A} = \begin{bmatrix}
    \frac{-C_f - C_r}{Mv} & \frac{C_r l_r - Mv^2 - C_f l_f}{Mv^2} & 0 & 0 & \frac{C_f}{Mvi_s} & 0 \\
    \frac{C_r l_r - C_f l_f}{J_z} & \frac{-C_r l_r^2 - C_f l_f^2}{J_z v} & 0 & 0 & \frac{C_f l_f}{J_z i_s} & 0 \\
    0 & 1 & 0 & 0 & 0 & 0 \\
    v & 0 & v & 0 & 0 & 0 \\
    0 & 0 & 0 & 0 & 0 & 1 \\
    0 & 0 & 0 & 0 & \frac{-C_s}{J_s} & \frac{-D_s}{J_s}
\end{bmatrix}
\]

where \(M\)(1600 kg)  is vehicle mass, \(J_z\)(1800 kg$\cdot$m$^2$) is yaw moment of inertia of the vehicle, \(v\)(120 km/hr) is the speed of the vehicle, \(l_f\)(0.9 m) and \(l_r\)(1.7 m) are distance from the vehicle’s center of gravity to the front and rear axles, respectively, \(C_f\)(45 kN/rad) and \(C_r\)(75 kN/rad) represents the cornering stiffness of the front and rear tires, and \(i_s\)(16) is steering ratio, relating the steering wheel angle to the tire angle. \(J_s\)(0.04 Nm$\cdot$s$^2$/rad) is inertia of the steering train, including the steering wheel and the driver’s neuromuscular system, \(C_s\)(1.1 Nm/rad) is stiffness of the steering train, and \(D_s\)(0.3 Nm$\cdot$s/rad) is damping coefficient of the steering train. The input vectors are given as:$
\mathbf{B_D} = 
        \bigg[ 0, 0, 0, 0, 0, \frac{1}{J_s} \bigg]^\top, 
\mathbf{B_A} =
        \bigg[ 0, 0, 0, 0, 0, \frac{1}{J_s} \bigg]^\top
$
\end{document}